\documentclass{www2009-submission}

\def\sharedaffiliation{
\begin{tabular}{c}
}
\usepackage{amsmath}

\begin{document}
\title{Alleviating Media Bias Through Intelligent Agent Blogging}
%
%

\numberofauthors{1}
%


\author{
%
\alignauthor Ernesto Diaz-Aviles\\
       \affaddr{L3S Research Center / Leibniz University Hannover}\\
       \affaddr{Appelstr. 4 D-30167. Hannover, Germany}\\
       \email{diaz@L3S.de}
}

\date{28 October 2008}

\maketitle

\begin{abstract}
Consumers of mass media must have a comprehensive, balanced and plural
selection of news to get an unbiased perspective; but achieving this goal can
be very challenging, laborious and time consuming. News stories development
over time, its (in)consistency, and different level of coverage across the
media outlets are challenges that a \textit{conscientious} reader has to
overcome in order to alleviate bias. In this paper we present an intelligent
agent framework currently facilitating analysis of the main sources of on-line
news in El Salvador. We show how prior tools of text analysis and Web 2.0
technologies can be combined with minimal manual intervention to help
individuals on their rational decision process, while holding media outlets
accountable for their work.
\end{abstract}
\small
%
%
%
%
%
\section{Introduction}
In a democratic system the news media plays an essential role: provide to the
public a comprehensive, timely and balanced collection of information, that
reflects the plurality of views for both individual and collective decisions.
The news media, however, is widely considered as biased~\cite{1999ASNE}.

El Salvador, a young democracy in Central America, does not scape to this issue.
Even though a more independent media is considered fundamental to the process of
democratisation~\cite{2006FUSADES}, the media companies with the biggest
circulation in the country are owned, primarily, by businessmen whose capital
comes from other economic activities which influence the media's content and
coverage. Journalists and analysts agree that popular mainstream media provides
information influenced by ideological, political and business bias that favours
the interests of the owners and
advertisers~\cite{2004FreedomOfExpressionInElSalvador, comunican2006}. On the
other hand, El Salvador also enjoys the benefits of the Web 2.0 paradigm shift.
Blogging has become very popular, Blogger.com\footnote{http://www.blogger.com/},
for example, ranks number nine on the sites of major traffic in the
country\footnote{http://www.alexa.com/}, above any newspaper electronic edition.
In this work, we aim to provide an intelligent agent based system that combines
prior tools for text analysis, with minimal manual processing, to facilitate
analysis of news media coverage over time.
	
The media can help individuals to make better decisions in their everyday lives.
Those decisions could pertain to health, safety, product selection, politics,
employment, personal finance, the environment, or other issues about which
individuals make purposive decisions. However, if the decision process is
rational, the consumers of mass media must have a comprehensive, balanced and
plural selection of news to get an unbiased perspective. Mullainathan and
Shleifer~\cite{SMAS05} present a theory of media bias based on the distribution
of preferences of readers. They showed that the crucial determinant of
information accuracy is not competition, per se, but consumer heterogeneity.
According to their economic model, on topics where reader beliefs diverge, e.g.,
political or financial issues, news providers slant and increase bias rather than
clear up confusion. Yet in the aggregate, a reader with access to all news
sources could get an unbiased view of the facts.

To achieve this unbiased perspective can be very challenging, laborious and time
consuming. News stories development over time, its (in)consistency, and different
level of coverage across the news outlets are challenges that a
\textit{conscientious} reader has to overcome in order to alleviate bias. We
present our first work toward this end, an intelligent agent framework currently
facilitating analysis of the main sources of on-line news in El Salvador.

\section{Intelligent Agent Blogging}
\begin{figure}
\centering
\includegraphics[width=0.6\columnwidth,clip=]{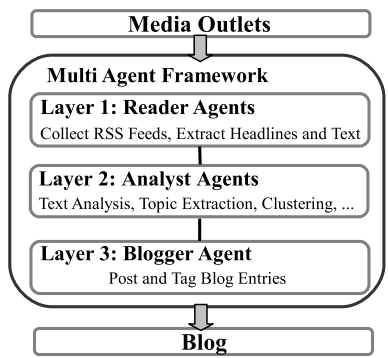}
\caption{Multi-Agent Framework}
\label{fig:arch}
\end{figure}

The architectural view of the multi-agent based framework consists of three
layers (Fig.~\ref{fig:arch}): ($i$) The first layer is composed by $reader$
$agents$, whose main task is to retrieve news articles from the predefined media
outlets. ($ii$) The media $analyst$ $agents$, in the second layer,
automatically organize (cluster) the digested news articles into thematic
categories~\cite{carrot2} and identify the latent $topics$ present on the
documents~\cite{2005ProbTopicModels}. Finally, based on the preprocessing of
the previous layer, ($iii$) The $blogger$ $agent$ on the third layer posts
related news articles summaries in a blog~\footnote{http://www.cuscatlan.org/},
tagging each entry of related issues with the topics identified, so as to
facilitate future search and retrieval.
\subsection{Data acquisition}
In order to analyze the different news offers, reader agents (one per media
outlet), first collect the headlines from the RSS feeds of three salvadorean
on-line newspapers\footnote{Media outlets were selected based on the circulation of their corresponding printed version, and also on the traffic rank reported by alexa.com}: 
\textit{La Prensa Gr\'{a}fica}\footnote{http://www.LaPrensaGrafica.com/},
\textit{El Diario de Hoy}\footnote{http://www.ElDiarioDeHoy.com/}, and
\textit{Diario Co Latino}\footnote{http://www.DiarioCoLatino.com/}. Diario
Co~Latino is considered left-leaning meanwhile the other two newspapers range
from right-leaning to ultra-right~\cite{2004FreedomOfExpressionInElSalvador}.
After collecting the headlines, reader agents parse the news articles' HTML
source, extracting their title and corresponding text. Then, they input this
data to the $analyst$ $agents$ responsible for text analysis.
\subsection{Text Analysis: key phrases, topics and clustering}
$Analyst$ $agents$ extract automatically organize small collections of
news articles into thematic categorie using $Carrot^2$~\cite{carrot2}, a 
search results clustering engine. The agents also extract the latent topics from
the texts using the state-of-the-art latent Dirichlet allocation (LDA)
algorithm implemented in MALLET~\cite{mallet}. Currently, one specialized agent
per task is used, but more can be instantiated to achieve a better degree of parallelization. The
results of the text analysis form the basis for the blog entries to be posted.
\subsection{Intelligent Agent Blogger}
The $Blogger$ $Agent$ is responsible for posting the summaries of the news
articles, where a single entry in the blog corresponds to the clusters found.
That is, each post contains the summaries of the news articles that are
related, with their corresponding link to the sources. The latent topics
identified are used as tags for the blog post, in order to create a more
coherent classification and to facilitate search. Here Blogger Data
API~\cite{bloggerAPI} was used for blog entry creation.
\section{Discussion}
The multi-agent framework, which in the end interface with the user through a
blog, is successful in aiding the discovery of interrelated news stories across
media outlets, facilitating the ideal scenario where a user is able to access
all sources of information to get an unbiased perspective. We opted to use a
blog because its format emphasizes a $participatory$ $journalism$, which in
this case Intelligent Agents play an active role in the process of collecting,
analyzing and disseminating information. Furthermore, a blog facilitates user
participation and discussion through comments, and it also helps to keep track
of news stories development over time.

Not surprisingly, the textual analysis tools are not optimized for the Spanish
language, and some special configuration, and small modifications were required
to achieve the desirable performance.

While several systems aim to provide a global aggregation of
news~\cite{googleNews}, and analysis of worldwide news stories
development~\cite{emm}, we focus on the niche of local news of El Salvador,
exploiting available tools of text analysis to cope with the bias exhibited by
the local mass media in their coverage. We consider our focus not a limitation,
but rather an answer to \textit{The Long Tail} phenomenon exhibited by those
services. 

Adapting the system to a different country or sources of information
would not involved a great effort, basically it would require the implementation
of new reader agents.

\section{Conclusions and Future Work}
We presented ongoing work towards alleviating media bias: an intelligent agent
based system for the analysis of news articles. We focus on a set of media
outlets from El Salvador, approximating the conditions of an ideal scenario where
a reader with access to all news sources could get an unbiased perspective. We
present a coherent organization of related news article within a blog, which is
automatically updated on a daily basis; easing the task for the users of finding
related information across multiple sources through a single location, keeping
track of the evolution of opinions and news development over time, e.g., using an
easily accessible blog archive, and allowing them to participate through comments
and discussions.

As future work, we plan to integrate sentiment analysis techniques to help us
determine the polarity of the news articles toward predefined entities, e.g.,
political parties, political figures or organizations. Besides that, we plan to
provide Personalized Recommendations of blog posts according to the reader's
particular interest. Furthermore, we plan to use our framework to also monitor
media partisan-bias during the coming elections in El Salvador, which will
provide us with a good opportunity to carry out a deeper user study.

Although our preliminary empirical evaluation is promising and support the
validity of our approach, which is also sustained by a solid economic model of
market for news~\cite{SMAS05}, we believe that our contribution is an initial
step on how the current Web 2.0 paradigm and readily available AI technology
impact processes of democratisation, e.g., providing the individuals with better
tools for their rational decisions, while holding media outlets accountable for
their work. Additional research in this field is still to be explored.


\section{Acknowledgments}
This work has been partially supported by the Programme Al$\beta$an, the European
Union Programme of High Level Scholarships for Latin America, scholarship
no(E07D400591SV).

%
\bibliographystyle{abbrv}
\begin{small}
	\bibliography{cadejo}  
\end{small}

%
%

\end{document}